# Salient Object Detection: A Distinctive Feature Integration Model

Abdullah J. Alzahrani, Hina Afridi

College of Computer Science and Engineering

*Abstract*— We propose a novel method for salient object detection in different images. Our method integrates spatial features for efficient and robust representation to capture meaningful information about the salient objects. We then train a conditional random field (CRF) using the integrated features. The trained CRF model is then used to detect salient objects during the online testing stage. We perform experiments on two standard datasets and compare the performance of our method with different reference methods. Our experiments show that our method outperforms the compared methods in terms of precision, recall, and F-Measure.

I. INTRODUCTION

Human eyes build on the ability to grab the interested regions accurately and rapidly in different scenes. This ability is called visual attention mechanism or saliency mechanism. Saliency is inferred from visual uniqueness, rarity, unpredictability, or surprise, which is often caused by variations of characteristics in images such as illumination, gradient, textures, color, boundaries, and edges.

Traditional saliency detection methods consider different saliency cues. One of them is contrast, which aims at analyzing the uniqueness of each image patch or image pixel with respect to local contexts or global ones. On one hand, these methods usually fail to preserve object details. On the other hand, they are often incapable to detect salient objects with large sizes. Itti et al. [1] computes image saliency using centre-surround differences over multiple scales image features. Ma et al. [2] evaluate saliency using color contrast in a neighborhood. Harel et al. [3] propose a graph based saliency method, which normalizes different feature maps by the balance distribution of Markov chains to present dominant regions. In [4][5], a hybrid model is proposed to detect salient regions. For this purpose, they combine local low-level cues [6][7][8], global features [9][10], visual organization rules [11], and high-level factors [12][13]. Xie et al. [14] consider local low-level features [15] and mid-level cues [16][17] within the Bayesian framework to detect saliency. Wang et al. [18] propose a bottom-up visual saliency detection algorithm.

Zhang et al. [19] propose a novel multi-scale spectral-spatial gradient based salient object detection method. Syed et al. [20] propose an efficient and robust method for face detection on a 3D point cloud represented by a weighted graph [21], Mu et al. [22] propose a novel deep neural network framework embedded with covariance descriptor [23] for salient object detection in low-contrast images [24]. Ren et al. [25] take color difference as the most important cue to detect salient objects. Peng et al. [26] propose a novel structured matrix decomposition method [27][28] with two regularizations: (1) a tree-structured sparsity-inducing regularization that absorbs the image structure [29] and enforces patches from the same object to have similar saliency information, and (2) a Laplacian regularization [30] that increases the gaps between salient objects and the background. Song et al. [31] propose a novel depth-aware salient object detection and segmentation framework [32][33][40] via multiscale discriminative saliency fusion and bootstrap learning for RGBD [34]. Hou et al. [35] present a deeply supervised salient object detection with short connections.

Li et al. [36] Propose a novel deep neural network framework embedded with low-level features (LCNN) [37][38] for salient object detection in complex images. Qu et al. [39] propose a new convolutional neural network (CNN) [41] to automatically learn the interaction mechanism for RGBD salient object detection. Chen et al. [42] presents a method for salient object detection. Their method exploits spatial and temporal cues as well as a local constraint to achieve a global saliency optimization. Liang et al. [43] extend the concept of salient object detection to material level based on hyperspectral imaging and propose a material-based salient object detection method which can distinguish objects with similar perceived color but different spectral responses. Xiao et al. [44] propose a saliency detection model with a combination of superpixel segmentation and eye tracking data. Wang et al. [45] propose a deep learning model [46] to efficiently detect salient regions in videos. Fu et al. [47] present a method, spectral salient object detection, that aims at maintaining objects holistically during pre-segmentation in order to provide more reliable feature extraction [48] from a complete object region and to facilitate object-level saliency estimation.

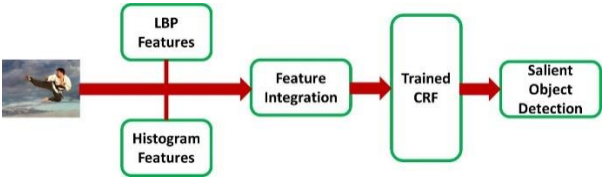

Figure 1. We combne LBP feature and histogram features y exploiting our integration model that transforms these features to a distinctive space. The trained CRF detect the salient object in an image.

To deal with the challenges of salient object detection, we propose a feature integration approach as shown in Fig. 1. Our method is based on the following observations:

(a) The accuracy of the saliency maps is sensitive to the orientation and size of regions as the salient objects may appear at different scales and orientations. The integration of different features can improve its robustness.

(b) Our integration method detects different objects regardless of their orientations and sizes. Moreover, our method uniformly highlights the contained regions.

(c) Considering huge training data comprising of different variations in terms of objects can improve the performance of saliency maps.

The contribution of proposed method is the integration model for combining spatial features to distinctively represent salient objects. Our method is composed of two major steps. Firstly, two different spatial features are integrated together to extract and formulate unique features. Secondly, these features are fed to the conditional random fields to learn the characteristics of salient object during the training stage and detect the salient objects during the testing stage. The experimental results demonstrate that our proposed method achieves very good performance in terms of precision, recall, and F-Measure.

The rest of the paper is organized as follows. Section 2 presents the feature integration model. The conditional random field (CRF) is presented in Section 3. In Section 4, the experimental results are presented and discussed. Finally, the paper is concluded in Section 5.

## II. FEATURE INTEGRATION MODEL

According to Singh et al. [49], the LBP operator is defined for the gray scale images. The general form of the operator for a circularly symmetric neighbor set of P members on a circle of radius R and center $(x_c, y_c)$, depicted by $LBP_{\{P, R\}} (x_c, y_c)$ is formulated:

$$LBP_{P,R}(x_c, y_c) = \sum_{p=0}^{P-1} S(I(x_p, y_p) - I(x_c, y_c)) \times 2^P$$

$$S(I(x_p,y_p) - I(x_c,y_c)) = \begin{cases} 1, & \text{if} \quad I(x_p, x_p) - I(x_c, y_c) \times 2^p \geq 0 \\ 0, & \text{otherwise} \end{cases}$$

$$C_{s,t} = \sqrt{u^2 + v^2}$$

In the equation, I(x, y) represents the intensity at pixel location (x, y) in the image I. For calculation of LBP, 8 neighborhood pixels P = 8 of a pixel at (x, y) are exploited. This is a faster approach to compute the LBP without considering a circular symmetry of the members which requires interpolation. The general case of the LBP presents P equally spaced locations on a circle of radius R and center $(x_c, y_c)$ whose coordinates are formulated as:

$$x_p = x_c + Rcos(2\pi p/P)$$
$$y_p = y_c + Rsin(2\pi p/P)$$

It is stated that p =0, 1, .........., P-1. The intensities that do not confine to the original positions are interpolated using bilinear interpolation. The parameter P computes the angular space between neighbors and the parameters R computes the size of the window. The purpose of a grid is to depict the locations of the neighbors which fall on the original positions of the pixels of the image I. It is worth noticing that the LBP operator of Singh et al. [49] is based on the concept of dividing or thresholding color pixels using a hyper-plane in m dimensions. Therefore, it is formulated as:

$$LBP(x_c, y_c) = \sum_{p=0}^{P-1} S(I_p) \times 2^P$$

$$S(I(x_p,y_p) - I(x_c,y_c)) = \begin{cases} 1, & E(I_p) \geq 0 \\ 0, & E(I_p) < 0 \end{cases}$$

Where

$$E(I_p) = E(r, g, b) = n_1(r_p - r_c) + n_2(g_p - g_c) + n_3(b_p - b_c)$$

The LBP operator considers texture patterns for a color image instead of considering gray scale images. It treats a color pixel as a vector having m components and creates a hyperplane. This hyperplane is exploited as a boundary to divide pixels into two classes. A color pixel in the neighborhood is assigned a value 1 if it lies on or above the plane, and the value 0, otherwise. Therefore, the LBP operator provides spatial relationship among color pixels, which represents local texture features for salient object in the scene.

Salient object patterns have much information including color cues and orientation components configurations. The spatial histogram features (SHF) [50] combine color cues and spatial structures. Histogram is a global representation of salient object patterns, which is translation invariant. We enlarge discrimination information in the histograms by two methods. First, we adopt color information as measurements of salient object patterns. Second, we introduce spatial histograms to extract salient object shape information. The spatial histogram is represented by the average spatial histogram of salient object samples which is formulated as:

$$SH_{m(i)}^{rect(x,y,w,h)} = \sum_{j=1}^{n} SH_{m(i)}^{rect(x,y,w,h)}(P_j)$$

where $P_j$ is an object sample, $m(i)$ is the measurement and $rect(x,y,w,h)$ is the spatial template. The histogram feature is defined as the distance to the average histogram formulated as:

$$f_{m(i)}^{rect(x,y,w,h)} = D(SH_{m(i)}^{rect(x,y,w,h)}(P_i), SH_{m(i)}^{rect(x,y,w,h)}(P_j))$$

We integrate the two feature as formulated:

$$IntegModel = \frac{1}{\sigma \times \mu}\left(\frac{LBP(x_c, y_c) \times f_{m(i)}^{rect(x,y,w,h)}}{2}\right)$$

Our feature integration model transforms the features to a distinctive space. Therefore, the integration represents high quality description of salient object in term of most distinctive information. The integration models the evolving relative spatial relationships among parts of any salient object present in a scene.

### III. CONDITIONAL RANDOM FIELD

The features we extracted are used to train conditional random field (CRF). Initially, a grid of particles is disposed on the image plane. Each particle represents a block of pixels of predefined size. The extracted features with the corresponding label sequence are used to learn the CRF parameters during the training stage, and the salient objects are inferred on the test samples. The collected features are latched to build a feature vector for each particle.

The goal of the feature filtering and identification is to isolate and filter out the features that would be possible to contribute less if considered singularly, but that they do contribute to the detection of salient objects if considered collectively.

A Conditional Random Field (CRF) is a discriminative technique exploited for labeling various structured and unstructured data. It mentions the probability of a particular label sequence, given the observation sequence. In general, x is our resultant sequence, comprising of N samples gathered within a sample set (i.e. $x = x_1, x_2, \ldots, x_N$), showing the extracted features. Provided the observation sequence, the CRF thus shows the most probable label in terms of dominant feature, computing the output label y ($y = y_1, y_2, \ldots, y_M$) of the respective salient object, as formulated in the equation:

$$p(y/\mathbf{x}; \mathbf{w}) = \frac{exp \sum_j w_j F_j(\mathbf{x}, y)}{Z(\mathbf{x}, \mathbf{w})}$$

In the equation, $F_j(x, y)$ is a feature function, which contains the mapping of input and output in the form of $F_j: X * Y \rightarrow R$. In the CRF, each feature function provides the score for any output label y in the form of its affiliation to the input observation feature x. The part of the equation that is the denominator in the above equation is a partition function $Z(x, w)$, which ranges over all the label set y.

$$Z(\mathbf{x}, \mathbf{w}) = \sum_{y'} exp\left\{\sum_j w_j F_j(\mathbf{x}, y')\right\}$$

Therefore, the function Z is working as a normalization element in the equation. Considering the extracted features x, the corresponding label is formulated as:

$$\hat{y} = argmax_y\, p(y/\mathbf{x}; \mathbf{w}) = argmax_y \sum_j w_j F_j(\mathbf{x}, y)$$

For each j, we computer various $F_j$ functions, depending on the parameter $w_j$ and the provided input x. The major target in getting the probabilistic score for each target label is that it is important to show the most probable feature and label for each observation, which can detect the salient object as the scene dynamically considering the entire dataset.

The objective of the training stage is to identify the proper magnitudes for the parameters $w_j$, so as to increase and maximize the conditional probability of the training samples. To handle this problem, we explore the stochastic gradient ascent to maximize the conditional log-likelihood (CLL) of the set of training samples, which is formulated as:

$$\frac{\partial}{\partial w_j} \log p(y/\mathbf{x}; \mathbf{w}) = F_j(\mathbf{x}, y) - \frac{\partial}{\partial w_j} \log Z(\mathbf{x}, \mathbf{w})$$

Considering each weight parameter $w_j$, the partial derivative of CLL is analyzed for single training samples, i.e., one weight for each feature function $F_j$. The partial derivative with respect to $w_j$ relates to the i-th magnitude of the feature function for its targeted label y, minus the averaged magnitudes of the feature function for all possible labels y. Therefore, the equation can be reformulated as:

$$\frac{\partial}{\partial w_j} \log p(y/\mathbf{x}; \mathbf{w}) = F_j(\mathbf{x}, y) - \sum_{y'} p(y'/\mathbf{x}; \mathbf{w})[F_j(\mathbf{x}, y')]$$

For the computation purpose, we need to maximize the conditional log-likelihood by stochastic gradient ascent, wj is updated according to the equation where alpha is the learning rate.

$$w_j = w_j + \alpha(F_j(\mathbf{x}, y) - \sum_{y'} p(y'/\mathbf{x}; \mathbf{w})[F_j(\mathbf{x}, y')])$$

## IV. EXPERIMENTAL RESULTS

We evaluate our method on two benchmark datasets namely MSRA-B [51] and SOD [52]. The MSRAB dataset consists of 5000 images and all images are labeled with pixel-wise ground truth [52]. This dataset is widely used for salient object detection. Most of the images comprise of only one salient object. The SOD dataset contains 300 images. Pixel-wise ground truth annotations for salient objects are contributed by Jiang et al. [52]. The images in SOD dataset consists of different salient objects of various sizes and positions. Five sample images from MSRA-B and SOD datasets are depicted in Fig. 2 and Fig. 3, respectively.

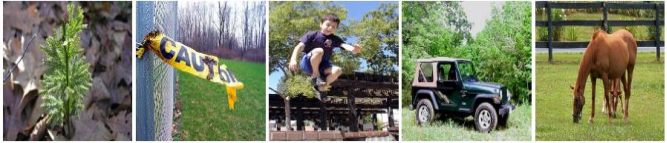

Figure 2. MSRA-B dataset. Five sample images from MSRA-B dataset are depicted where salient objects are localized in the center of the images.

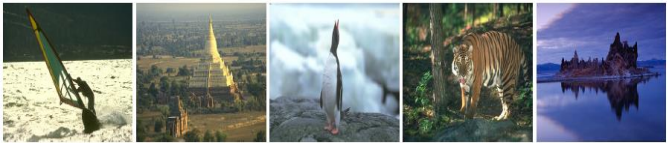

Figure 3. SOD dataset. Five sample images from SOD dataset are depicted where salient objects are localized in the center of the images.

We present quantitative results of our method on the two datasets and compare them with SIFT features [53], LBP features [49] only, and SHF [50] features only. We consider three performance metrics namely precision, recall, and F-Measure to extensively evaluate the experimental results. The results in terms of precision, recall, and F-Measure for the reference methods and our proposed method are presented in Table 1. Our proposed method outperforms all the reference methods considering the three performance metrics. For the individual features SIFT, LBP, and temp, we used the same set of training data to maintain consistency with our model.

Table 1. Precision. The precision of our proposed method is higher than other reference methods. Therefore, our method outperforms in salient object detection considering the two datasets.

| Datasets | SIFT | LBP | SHF | Our |
|---|---|---|---|---|
| MSRA-B | 0.48 | 0.53 | 0.58 | 0.78 |
| SOD | 0.51 | 0.57 | 0.54 | 0.77 |

## V. CONCLUSION

We proposed a novel method for salient object detection in images. Our method represented high quality description of salient objects in term of most distinctive information. The performance of our proposed method is evaluated on two datasets and compared to different reference methods. The performance metrics precision, recall, and F-Measure show that our method outperformed the reference methods.

As a future work, we would also extend our proposed method to detect various other salient objects by considering more datasets.